\documentclass[onecolumn]{article}
\usepackage[margin=1in]{geometry}
\usepackage{booktabs}
\usepackage{graphicx} 
\usepackage{rotating} 
\usepackage{multirow}
\usepackage{makecell}
\usepackage{amsmath}
\usepackage{authblk}            
\usepackage{url}

\usepackage{array} 
\newcolumntype{C}[1]{>{\centering\arraybackslash}m{#1}}

\newcommand\blfootnote[1]{%
  \begingroup
  \renewcommand\thefootnote{}%
  \footnotetext{#1}%
  \endgroup
}

\title{Beyond Aggregate Scores: Behavioral Correctness Assumptions for Assessing Reference-Based Automatic Evaluation Methods}

\author[1]{Maria Mahbub\thanks{Corresponding author: \texttt{mahbubm@ornl.gov}}}
\author[1]{Ashley Rice}
\author[2]{Michael R. Munroe}
\author[2]{Amidu Kamara}
\author[1]{Amir Sadovnik}

\affil[1]{Oak Ridge National Laboratory, Oak Ridge, TN 37830, USA}
\affil[2]{DHS Science and Technology Directorate, Washington, DC 20528, USA}

\date{} 

\begin{document}

\maketitle

\begin{abstract}
Automated reference-based evaluation methods play a critical role in assessing natural language generation systems. Existing meta-evaluation primarily measures agreement with human judgments or benchmark labels, providing limited insight into evaluator behavior under controlled conditions. We introduce behavioral correctness assumptions, a complementary framework for evaluating reference-based automatic evaluation methods. We define a taxonomy of correctness-preserving and correctness-altering assumptions and operationalize them through controlled response transformations that specify expected scoring behaviors. We evaluate diverse lexical, character-level, semantic, LLM-based, and hybrid evaluators and analyze their assumption-level behavior, stability, sensitivity, repeat-run variability, configuration sensitivity, and reproducibility. Our experiments reveal distinct behavioral trade-offs across evaluation paradigms: no evaluator satisfies all proposed correctness assumptions, and evaluators with similar aggregate performance can exhibit substantially different behavioral profiles. These findings demonstrate that behavioral correctness assumptions provide diagnostic information obscured by conventional aggregate meta-evaluation.
\end{abstract}

\section{Introduction}
\blfootnote{Notice: This manuscript has been authored by UT-Battelle, LLC, under contract DE-AC05-00OR22725 with the US Department of Energy (DOE). The US government retains and the publisher, by accepting the article for publication, acknowledges that the US government retains a nonexclusive, paid-up, irrevocable, worldwide license to publish or reproduce the published form of this manuscript, or allow others to do so, for US government purposes. DOE will provide public access to these results of federally sponsored research in accordance with the DOE Public Access Plan (\url{https://www.energy.gov/doe-public-access-plan}).}

As large language models (LLMs) become more capable and are adopted in high-stakes applications, reliable automatic evaluation has become as important as model generation itself. In tasks such as question answering, summarization, retrieval-augmented generation (RAG), and conversational AI, the quality of LLM outputs is routinely assessed by comparing generated responses against one or more reference answers using evaluation methods. These methods support model benchmarking, system optimization, and increasingly serve as training objectives and deployment criteria. 

Despite their widespread adoption, evaluation methods frequently produce conflicting judgments. For the same response, a lexical overlap metric may assign a low score due to surface-level differences, while a semantic similarity metric assigns a much higher score. LLM-based evaluators may further disagree depending on the evaluator model, prompting strategy, or the formulation of the answer. Such discrepancies have become increasingly common as evaluation has evolved from token matching toward semantic and reasoning-based approaches.

Existing works primarily assess automatic evaluation methods using aggregate measures such as benchmark performance, agreement with human judgments, or meta-evaluation benchmarks designed specifically for LLM judges \cite{li2024llms, chang2024survey, li2025generation, gao2025llm}, providing limited insight into \emph{how} these methods respond to specific variations in the evaluated response. In particular, they do not reveal whether an evaluation method responds appropriately to common correctness-preserving and correctness-altering variations, such as paraphrasing, adding correct information, omitting relevant facts, introducing hallucinations, or violating logical consistency.

To this end, we propose a diagnostic framework for assessing reference-based evaluation methods using explicit behavioral correctness assumptions that specify how reliable evaluators should respond to controlled variations in response quality. We define a taxonomy of these assumptions and operationalize each through a controlled response transformation designed to isolate a single aspect of response quality while minimizing confounding effects. Together, these transformations form a test suite that systematically evaluates whether an evaluation method satisfies each assumption. Rather than introducing another evaluation metric or benchmark, our framework provides a methodology for characterizing the behavior of reference-based evaluation methods. Beyond the primary behavioral analysis, we investigate repeat-run variability, evaluator configuration sensitivity, and test-suite reproducibility, demonstrating that the proposed framework produces consistent evaluator characterizations across repeated evaluations, decoding configurations, and independently generated test suites. Our contributions are summarized as follows:

\begin{itemize}
    \item We introduce a diagnostic framework for assessing reference-based evaluation methods using a taxonomy of behavioral correctness assumptions that characterize desirable properties of reliable evaluators.

    \item We develop and validate a transformation-based test suite that operationalizes each correctness assumption through controlled response transformations.

    \item We conduct a comprehensive empirical study spanning lexical, character-level, semantic, LLM-based, and hybrid evaluation methods, demonstrating that assumption-level behavioral analysis reveals systematic differences that are largely obscured by conventional aggregate benchmark evaluation.
\end{itemize}

\section{Related Work}
\subsection{Reference-based automatic evaluation} Traditional reference-based evaluation methods, including BLEU \cite{papineni2002bleu}, ROUGE \cite{lin2004rouge}, METEOR \cite{banerjee2005meteor}, and chrF \cite{popovic2015chrf}, rely primarily on lexical and character-level overlap between generated responses and reference answers. Semantic and learned metrics, such as MoverScore \cite{zhao2019moverscore}, BERTScore \cite{zhang2019bertscore}, BLEURT \cite{sellam2020bleurt}, and BARTScore \cite{yuan2021bartscore}, extend this paradigm beyond surface overlap using contextual representations or learned models. While these methods differ substantially in formulation and their agreement with human judgments \cite{hu2024unveiling,peng2024survey}, conventional evaluations provide limited insight into how individual metrics respond to controlled changes in response correctness.

\subsection{LLM-based evaluation and meta-evaluation} LLM-based evaluators assess responses through scoring, ranking, classification, or natural-language critique. Approaches such as G-Eval \cite{liu2023g} and specialized judge models such as Prometheus \cite{kim2023prometheus}, PandaLM \cite{wang2024pandalm}, and JudgeLM \cite{zhu2025judgelm} demonstrate the effectiveness of LLMs as automatic evaluators. However, LLM-based evaluation is sensitive to factors including prompting, evaluator model, position and presentation biases, and inference configuration \cite{li2024llms,gu2026survey}. Meta-evaluation therefore typically assesses evaluators through correlation with human judgments, pairwise agreement, ranking accuracy, or dedicated benchmarks such as JudgeBench \cite{tan2025judgebench}. These approaches quantify overall evaluator performance but provide limited insight into why different evaluation methods disagree or which behavioral properties are responsible for those differences.

\subsection{Behavioral testing} Behavioral testing provides a complementary perspective by evaluating systems under controlled changes designed to probe specific expected behaviors. CheckList \cite{ribeiro2020beyond}, for example, introduced invariance and directional expectation tests for diagnosing NLP models beyond aggregate held-out accuracy. Related work has applied controlled perturbations to LLM-based semantic similarity judges to characterize their sensitivity to factors such as perturbation type, position, and context \cite{aksoy2026semantic}. Our work extends this perspective to reference-based automatic evaluation by defining explicit behavioral correctness assumptions and testing whether evaluators respond appropriately to correctness-preserving and correctness-altering transformations. Rather than measuring perturbation sensitivity alone, the proposed framework specifies the expected direction of evaluator behavior, providing an assumption-level diagnostic characterization that complements conventional benchmark-based meta-evaluation.

\section{Methodology}

Traditional evaluations of reference-based automatic evaluation methods primarily rely on aggregate measures such as benchmark performance, agreement with human judgments, or correlation with other evaluation methods. While these measures provide useful estimates of overall performance, they do not assess whether an evaluation method behaves as expected under specific changes to a generated response. Consequently, two evaluation methods may achieve similar aggregate performance while exhibiting substantially different behavior when evaluating paraphrases, incomplete responses, hallucinations, or alternative correct answers.

\begin{figure*}[!ht]
    \centering
    \includegraphics[width=\textwidth]{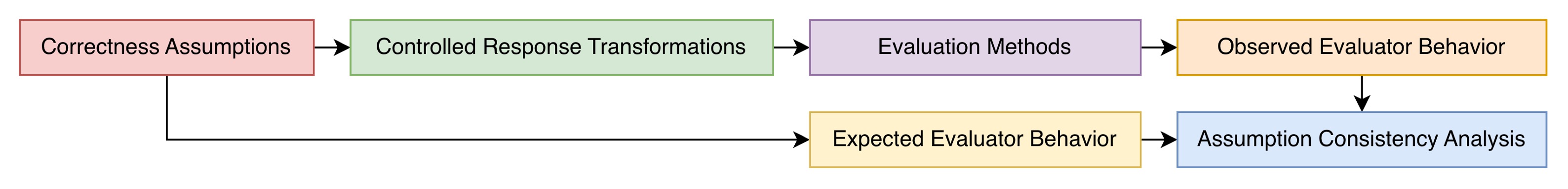}
    \caption{Overview of the behavioral correctness framework for assessing reference-based evaluation methods.}
    \label{fig:overview}
\end{figure*}

We propose a complementary evaluation methodology based on explicit correctness assumptions that characterize desirable properties of reliable reference-based evaluators. As illustrated in Figure~\ref{fig:overview}, each assumption is operationalized through controlled response transformations that specify expected evaluator behavior under a particular change in response quality. Comparing the expected and observed scoring behavior provides a fine-grained characterization of evaluator strengths and limitations beyond aggregate performance.

\subsection{Correctness Assumptions}
\label{sec:assumptions}

We characterize evaluator reliability through a set of correctness assumptions that specify expected scoring behavior under controlled response transformations. Rather than treating evaluation quality as a single aggregate quantity, these assumptions enable different aspects of evaluator behavior to be tested independently.

We organize the proposed correctness assumptions into two categories based on the expected effect of the corresponding response transformation.
\textit{Correctness-preserving assumptions} describe transformations that change wording, structure, style, or other characteristics without altering correctness with respect to the reference. Evaluators should therefore assign similar scores before and after these transformations.
\textit{Correctness-altering assumptions} describe transformations that modify response quality, such as introducing factual errors, omitting relevant information, or adding irrelevant content. Reliable evaluators should therefore be sensitive to these changes and assign lower scores.

Table~\ref{tab:correctness_assumptions_perturbations} summarizes the proposed assumptions and corresponding transformations. Although this taxonomy is not intended to be exhaustive, it captures a broad set of properties commonly encountered in reference-based evaluation and provides a principled basis for systematically assessing evaluators.

\begin{table*}[!htbp]
\centering
\caption{Correctness assumptions and controlled response transformations used to evaluate desirable properties of reliable reference-based automatic evaluation methods. Expected $\Delta$ denotes the expected change in evaluator score after applying the transformation relative to the original response.}
\label{tab:correctness_assumptions_perturbations}
\begin{tabular}{p{2.8cm} p{5.2cm} p{4.3cm} p{2cm}}
\toprule
\textbf{Category} &
\textbf{Correctness Assumption} &
\textbf{Transformation} &
\textbf{Expected $\Delta$} \\
\midrule

\multirow{8}{*}{\makecell{Correctness-\\Preserving}}
& Semantic Invariance
& Paraphrase
& $\approx 0$ \\

\cmidrule{2-4}

& \multirow{2}{*}{Length Invariance}
& Verbose
& $\approx 0$ \\

&
& Concise
& $\approx 0$ \\

\cmidrule{2-4}

& Structural Invariance
& Structure Change
& $\approx 0$ \\

\cmidrule{2-4}

& \multirow{2}{*}{Style Invariance}
& Uncertain Style
& $\approx 0$ \\

&
& Certain Style
& $\approx 0$ \\

\cmidrule{2-4}

& Alternative Answer Recognition
& Alternative Correct Answer
& $\approx 0$ \\

\cmidrule{2-4}

& Additional Information Tolerance
& Correct Fact Added
& $\approx 0$ \\

\midrule

\multirow{6}{*}{\makecell{Correctness-\\Altering}}
& Recall Sensitivity
& Key Detail Omitted
& $< 0$ \\

\cmidrule{2-4}

& Precision Sensitivity
& Incorrect Fact Added
& $< 0$ \\

\cmidrule{2-4}

& Fine-Grained Factual Sensitivity
& Subtle Factual Error
& $< 0$ \\

\cmidrule{2-4}

& Logical Consistency
& Logical Contradiction
& $< 0$ \\

\cmidrule{2-4}

& Relevance Sensitivity
& Irrelevant Sentence Added
& $< 0$ \\

\cmidrule{2-4}

& Partial Correctness Sensitivity
& Partially Correct
& $< 0$ \\

\bottomrule
\end{tabular}
\end{table*}

\subsection{Controlled Response Transformations}
\label{sec:transformations}
Each correctness assumption is operationalized through one or more controlled response transformations. Starting from a baseline response, we modify the targeted property while keeping other characteristics as unchanged as possible, allowing changes in evaluator scores to be attributed to the transformation rather than unrelated response variation. Transformations are designed to isolate one correctness property at a time.
For each baseline response, we instantiate the transformations in Table~\ref{tab:correctness_assumptions_perturbations}, producing a transformation-based test suite for systematically probing evaluator behavior. The test-suite construction procedure is described in Section~\ref{sec:test_suite}.

\subsection{Evaluation Methods}
\label{sec:metrics}
The proposed framework is independent of any particular evaluation method and can be applied to any reference-based automatic evaluator. To demonstrate its generality, we evaluate a diverse set of widely used evaluation methods spanning traditional deterministic metrics and recent LLM-based evaluators.
Specifically, we evaluate lexical, character-level, semantic, hybrid, and LLM-based evaluators. The first three are deterministic, whereas hybrid and LLM-based evaluators may be stochastic.
More details on evaluation methods are provided in Appendix \ref{app:llm_config}.

\subsection{Assumption Satisfaction Analysis}
\label{sec:analysis}
We characterize evaluator behavior at three levels: individual transformations, correctness assumptions, and overall stability--sensitivity.
Let $R_0$ denote a baseline response, $R_t$ denote its transformed response obtained by applying transformation $t$, and $M(G,R)\in[0,1]$ is the score assigned by evaluation method $M$ when comparing response $R$ against reference answer $G$.

\paragraph{Transformation-level analysis}
For each evaluation instance, we compute the score difference between the transformed response and its corresponding baseline response as \[\Delta(R_t) = M(G,R_t)-M(G,R_0)\].
For a transformation type $t$, let $\mathcal{D}_t$ denote the set of evaluation instances generated using transformation $t$. The average transformation-level score difference is computed as
\[\Delta_t(M) = \frac{1}{|\mathcal{D}_t|} \sum_{R_t\in\mathcal{D}_t} \Delta(R_t)\]
Thus, $\Delta_t(M) \approx 0$ is expected for correctness-preserving transformations, whereas $\Delta_t(M) < 0$ is expected for correctness-altering transformations.
The transformation-level score differences provide a fine-grained characterization of evaluator behavior.

\paragraph{Assumption-level analysis}
Because some correctness assumptions are operationalized using multiple transformations, we also aggregate transformation results at the assumption level. For each correctness assumption $a$, let $\mathcal{T}_a$ denote the set of transformations associated with that assumption. The assumption-level score difference is computed as
\[\Delta_a(M) = \frac{1}{|\mathcal{T}_a|} \sum_{t\in \mathcal{T}_a} \Delta_t(M)\]
This aggregation yields one score difference per correctness assumption and evaluation method. For assumptions that are operationalized by a single transformation, the assumption-level score is equivalent to the corresponding transformation-level score. The assumption-level score differences therefore summarize evaluator behavior with respect to the proposed correctness assumptions rather than individual transformation implementations.

\paragraph{Summary-level analysis}
We summarize evaluator behavior using two complementary quantities: \textit{stability}, which measures the ability to produce consistent scores when the response remains equally correct despite superficial modifications under correctness-preserving transformations, and \textit{sensitivity}, which measures the ability to detect genuine degradations in response quality under correctness-altering transformations.

For each evaluation instance corresponding to a correctness-preserving transformation, stability is computed as
\[S_{\mathrm{stab}}(M;R_t) = max\left(0, 1 - \frac{\left|M(G,R_t) - M(G,R_0)\right|}{M(G,R_0)+\epsilon}\right)\], while for each evaluation instance corresponding to a correctness-altering transformation, sensitivity is computed as
\[S_{\mathrm{sens}}(M;R_t) = \frac{\max\left(0, M(G,R_0) - M(G,R_t)\right)}{M(G,R_0)+\epsilon}\]

Let $\mathcal{P}_{\mathrm{stab}}$ and $\mathcal{P}_{\mathrm{sens}}$ denote the set of valid baseline--transformed response pairs generated from correctness-preserving and correctness-altering transformations, respectively.
Overall stability and sensitivity scores for evaluation method $M$ are
\[\mathrm{Stability}(M) = \frac{1}{|\mathcal{P}_{\mathrm{stab}}|} \sum_{\mathcal{P}_{\mathrm{stab}}} S_{\mathrm{stab}}(M;R_t)\] and
\[\mathrm{Sensitivity}(M) = \frac{1}{|\mathcal{P}_{\mathrm{sens}}|} \sum_{\mathcal{P}_{\mathrm{sens}}} S_{\mathrm{sens}}(M;R_t)\]

Together, these analyses provide complementary characterizations of evaluator behavior at transformation, assumption, and summary levels.

\subsection{Evaluation Test Suite}
\label{sec:test_suite}
We construct the evaluation test suite in three stages: reference answer construction, baseline response generation, and controlled response transformation. Starting from source documents, we generate question--reference pairs, produce a natural baseline response by rewriting each reference answer, and apply controlled transformations corresponding to the correctness assumptions.

\paragraph{Reference Answer Generation}
The source corpus consists of nine PDF documents, detailing procedures and operations for the U.S. Coast Guard. They provide the factual basis for all reference answers and response transformations used throughout the evaluation. We extract text from the documents using OCR and use it as context for Llama-3.3-70B-Instruct \cite{grattafiori2024llama} to generate question--reference pairs. Generation prompts and additional details are provided in Appendix~\ref{app:prompt_reference}.

\paragraph{Transformation Generation}
Rather than transforming reference answers directly, we generate a natural baseline response for each question by prompting an LLM to rewrite the reference while preserving its semantic content. We then apply the controlled transformations defined in Section~\ref{sec:transformations} to each baseline response. Generation prompts and implementation details are provided in Appendix~\ref{app:prompt_transformation}.

\paragraph{Quality Control}
A human reviewer inspected 10 randomly selected samples per transformation to verify that each instance reflected its intended correctness assumption. For correctness-preserving transformations, the reviewer verified that correctness was retained and only the targeted characteristic changed; for correctness-altering transformations, the reviewer verified that the intended violation was introduced without unrelated changes. Invalid instances were revised or regenerated before inclusion.

The final test suite contains 203 question--reference pairs derived from nine source documents and 2,842 transformed responses, yielding 3,045 total evaluation instances. Response-length statistics across references, baselines, and transformation types are reported in Appendix \ref{app:data_statistics}.

\subsection{Robustness Analyses}
\label{sec:robustness}

\paragraph{Repeat-run Variability}
To assess stochastic variability in LLM-based evaluation, we repeat the complete evaluation procedure five times for two representative LLM-based and hybrid evaluators under identical settings. This analysis measures whether the observed behavior is reproducible across repeated runs of the same evaluator. For each correctness assumption, we compute the average score for each run across all evaluation instances associated with that assumption. We then quantify repeat-run variability for each assumption by computing the standard deviation across these run-level means. Lower standard deviation indicates greater consistency across repeated executions.

\paragraph{Configuration Sensitivity}
We assess sensitivity to evaluator backbone and decoding configuration using Prometheus-2 (8$\times$7B) \cite{kim2024prometheus}, Llama-3.1-8B \cite{grattafiori2024llama}, and Qwen-3-8B \cite{yang2025qwen3} for two representative LLM-based and hybrid evaluators.
For each model, we vary one decoding parameter at a time: temperature $\{0.0,0.2,0.5\}$, top-$p$ $\{0.8,0.9,1.0\}$, and top-$k$ $\{20,40\}$, holding the remaining parameters at their default values.
We summarize configuration sensitivity using the mean and standard deviation of assumption-level scores across configurations for each language model. Lower variability would suggest that evaluator behavior is not sensitive to the underlying language model or decoding strategy. Comparing these score distributions across correctness assumptions would further reveal whether certain ones are more susceptible to configuration changes than others.

\paragraph{Test Suite Reproducibility}

Because the test suite relies on LLM-generated baselines and transformations, we assess whether its resulting evaluator characterizations depend on the generator.
Using the same questions, reference answers, retrieved context, prompting strategy, and decoding configuration, we independently regenerate a subset of the test suite with a second language model.
We apply the same evaluators and compute assumption-level score differences following Section~\ref{sec:analysis}.

For evaluation method $M$, we quantify test suite reproducibility using the mean absolute difference between the assumption-level profiles obtained from the two independently generated test suites:
\[\mathrm{MAE}_{\mathrm{assump}}(M)
=
\frac{1}{|\mathcal{A}|}
\sum_{a\in\mathcal{A}}
\left|
\Delta_{a}^{(A)}(M)
-
\Delta_{a}^{(B)}(M)
\right|\],
where \(\mathcal{A}\) denotes the set of correctness assumptions, and $\Delta_{a}^{(A)}(M)$ and $\Delta_{a}^{(B)}(M)$ denote the assumption-level score differences for generators $A$ and $B$, respectively. Lower values would indicate that regenerating the test suite has little effect on the assumption-level characterization of evaluation method \(M\).

\section{Results}
In this section, we present the results of the proposed diagnostic framework.

\subsection{Transformation-Level Analysis}
\begin{figure}[!htbp]
    \centering
    \includegraphics[width=\linewidth]{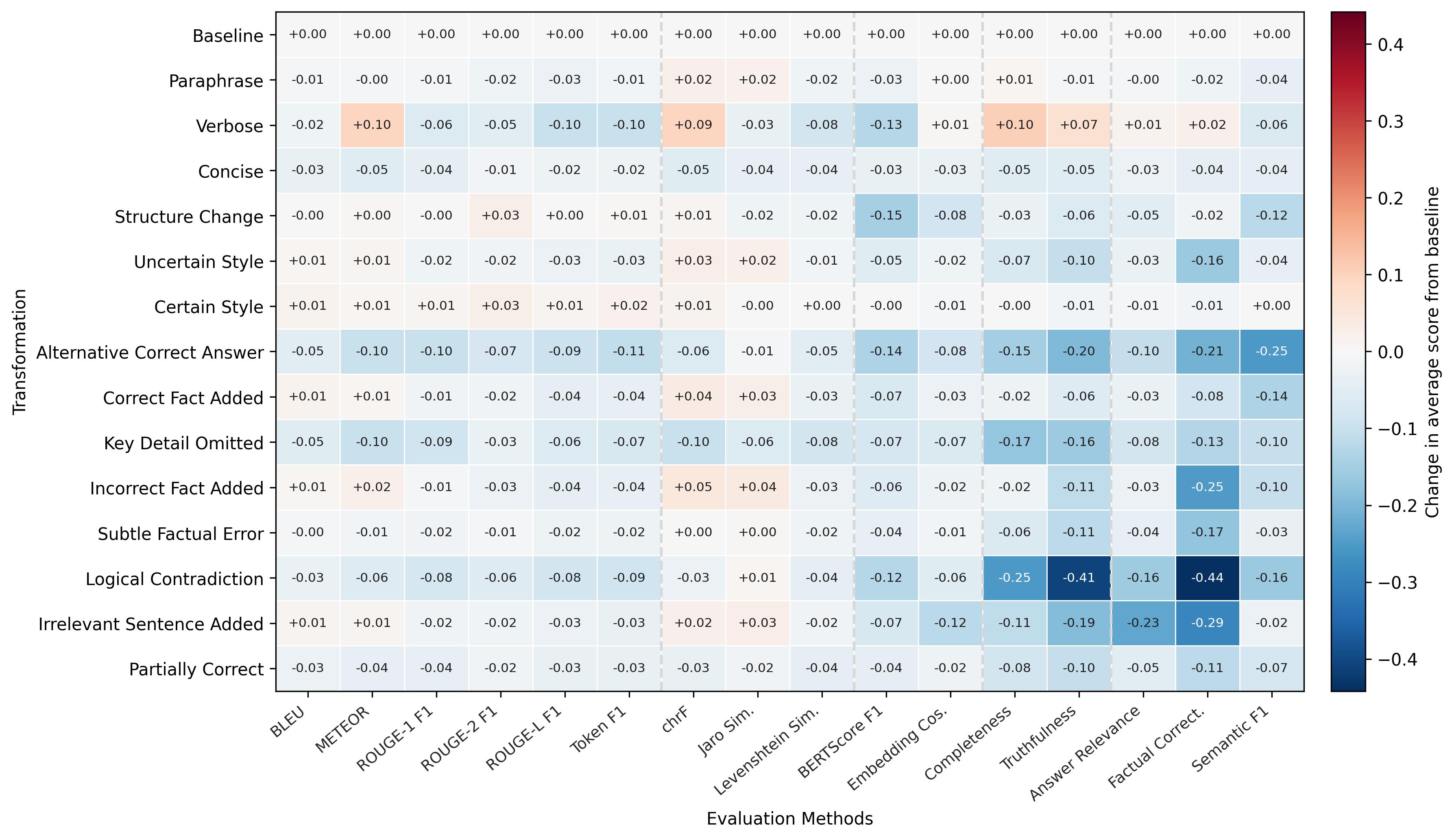}
    \caption{Average score difference relative to the baseline response for each controlled transformation and evaluation method. Positive values (red) indicate that the transformed response receives a higher score than the corresponding baseline response, while negative values (blue) indicate a lower score.}
    \label{fig:transformation_level}
\end{figure}

Figure \ref{fig:transformation_level} shows the average score difference relative to the corresponding baseline for each controlled transformation and evaluator; corresponding absolute scores are reported in Appendix \ref{app:results_trans}.

Lexical, character-based, and semantic similarity metrics generally exhibit smaller score changes across transformations than LLM-based and hybrid evaluators, consistent with greater invariance to the specific type of response modification. However, notable exceptions occur; for example, BERTScore decreases substantially under \emph{Verbose} (-0.13), \emph{Structure Change} (-0.15) and \emph{Alternative Correct Answer} (-0.14), while Embedding Cosine Similarity decreases under \emph{Irrelevant Sentence Added} (-0.12).

Among correctness-preserving transformations, \emph{Alternative Correct Answer} is particularly challenging: nearly every evaluator assigns lower scores to alternative valid responses. The effect extends across evaluator paradigms and is especially pronounced for Semantic F1 (-0.25), Factual Correctness (-0.21), and Truthfulness (-0.20), indicating that recognizing alternative valid answers remains challenging even for hybrid and LLM-based evaluators.

Correctness-altering transformations produce more heterogeneous responses among LLM-based and hybrid evaluators. \emph{Logical Contradiction}, for example, produces substantially larger reductions for Truthfulness (-0.41) and Factual Correctness (-0.44) than for Completeness (-0.25) or Answer Relevance (-0.16). Similar variation occurs for other violations: \emph{Incorrect Fact Added}, for instance, decreases Factual Correctness by 0.25 but produces little change for several other evaluators. These differences indicate that LLM-based evaluators emphasize different aspects of response quality.
Moreover, the absolute score distributions, demonstrated in Appendix \ref{app:results_trans}, indicate that even when substantial score reductions occur, transformed responses often continue to receive moderately high scores, suggesting that several evaluators detect correctness violations without penalizing them sufficiently.

The \emph{Verbose} transformation reveals a different failure mode. Despite preserving correctness, METEOR (+0.10), chrF (+0.09), Completeness (+0.10), and Truthfulness (+0.07) assign higher scores to verbose responses, suggesting a preference for additional lexical content or supporting information. In contrast, several similarity-based methods penalize verbosity, including BERTScore (-0.13), ROUGE-L (-0.10), and Token-F1 (-0.10).

Overall, the transformation-level results reveal two complementary failure modes exhibited by existing evaluation methods. Some evaluators incorrectly penalize transformations that preserve correctness, while others fail to sufficiently penalize transformations that introduce genuine correctness violations. Rather than identifying a universally reliable evaluation method, the transformation-level analysis demonstrates that each evaluator exhibits distinct strengths and weaknesses with respect to different correctness assumptions.

\subsection{Assumption-Level Analysis}
\begin{figure}[!htbp]
    \centering
    \includegraphics[width=\linewidth]{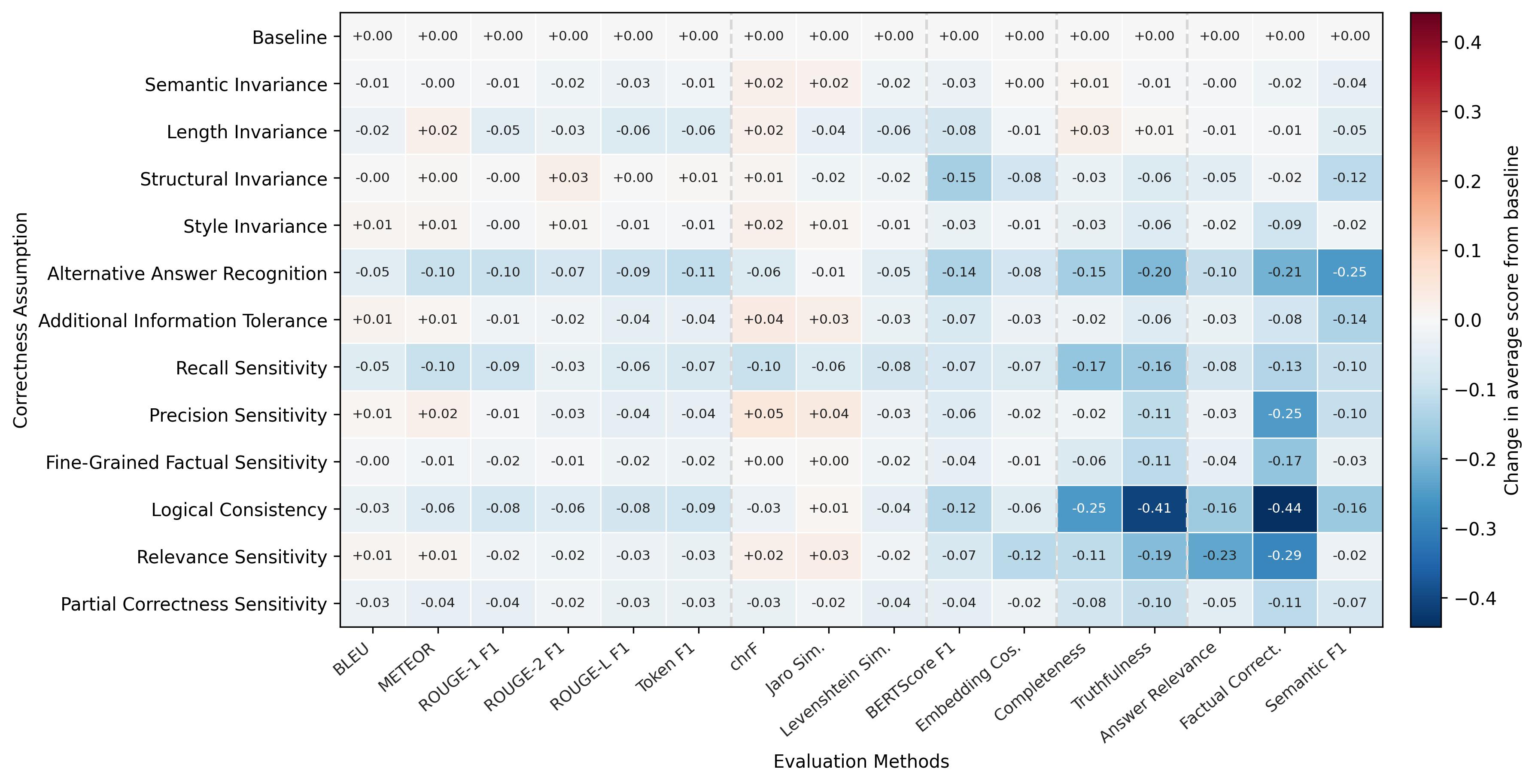}
    \caption{Average score difference relative to the baseline response for each correctness assumption and evaluation method. Positive values (red) indicate that the transformed response receives a higher score than the corresponding baseline response, while negative values (blue) indicate a lower score.}
    \label{fig:assumption_level}
\end{figure}

Figure~\ref{fig:assumption_level} aggregates transformation-level score differences by correctness assumption, providing an assumption-level behavioral profile for each evaluator.

Among correctness-preserving assumptions, \emph{Alternative Answer Recognition} is the most challenging across evaluator families, whereas \emph{Semantic Invariance} and \emph{Style Invariance} are generally well satisfied. The remaining assumptions reveal more evaluator-specific behavior; for example, \emph{Structural Invariance} exposes increased sensitivity for several semantic similarity and hybrid evaluators such as BERTScore (-0.15), Embedding cosine similarity (-0.08) and Semantic F1 (-0.12), while \emph{Additional Information Tolerance} indicates that a few methods continue to penalize responses containing correct supplementary information such as Semantic F1 (-0.14).

Correctness-altering assumptions exhibit greater variation across evaluators. \emph{Logical Consistency} produces the strongest responses overall, particularly for Truthfulness (-0.41) and Factual Correctness (-0.44). Other assumptions reveal more specialized sensitivities: Factual Correctness responds strongly to \emph{Precision Sensitivity} (-0.25) and \emph{Relevance Sensitivity} (-0.29), while Completeness and Truthfulness are more responsive to \emph{Recall Sensitivity} (-0.17 and -0.16, respectively). These differences indicate that LLM-based evaluators emphasize distinct dimensions of response quality rather than a uniform notion of correctness.

The aggregated assumption profiles also reveal a fundamental distinction between evaluation paradigms. Similarity-based evaluation methods exhibit relatively uniform behavior across the correctness assumptions, reflecting their reliance on overall similarity between the response and reference. In contrast, LLM-based and hybrid evaluators display substantially more differentiated assumption profiles, indicating that they encode richer notions of response quality. However, this increased discrimination does not necessarily correspond to improved evaluator behavior, as several correctness-preserving assumptions continue to receive unnecessary penalties while some correctness-altering assumptions remain only weakly penalized.

\subsection{Summary-Level Analysis}
\begin{figure}[!htbp]
    \centering
    \includegraphics[width=\linewidth]{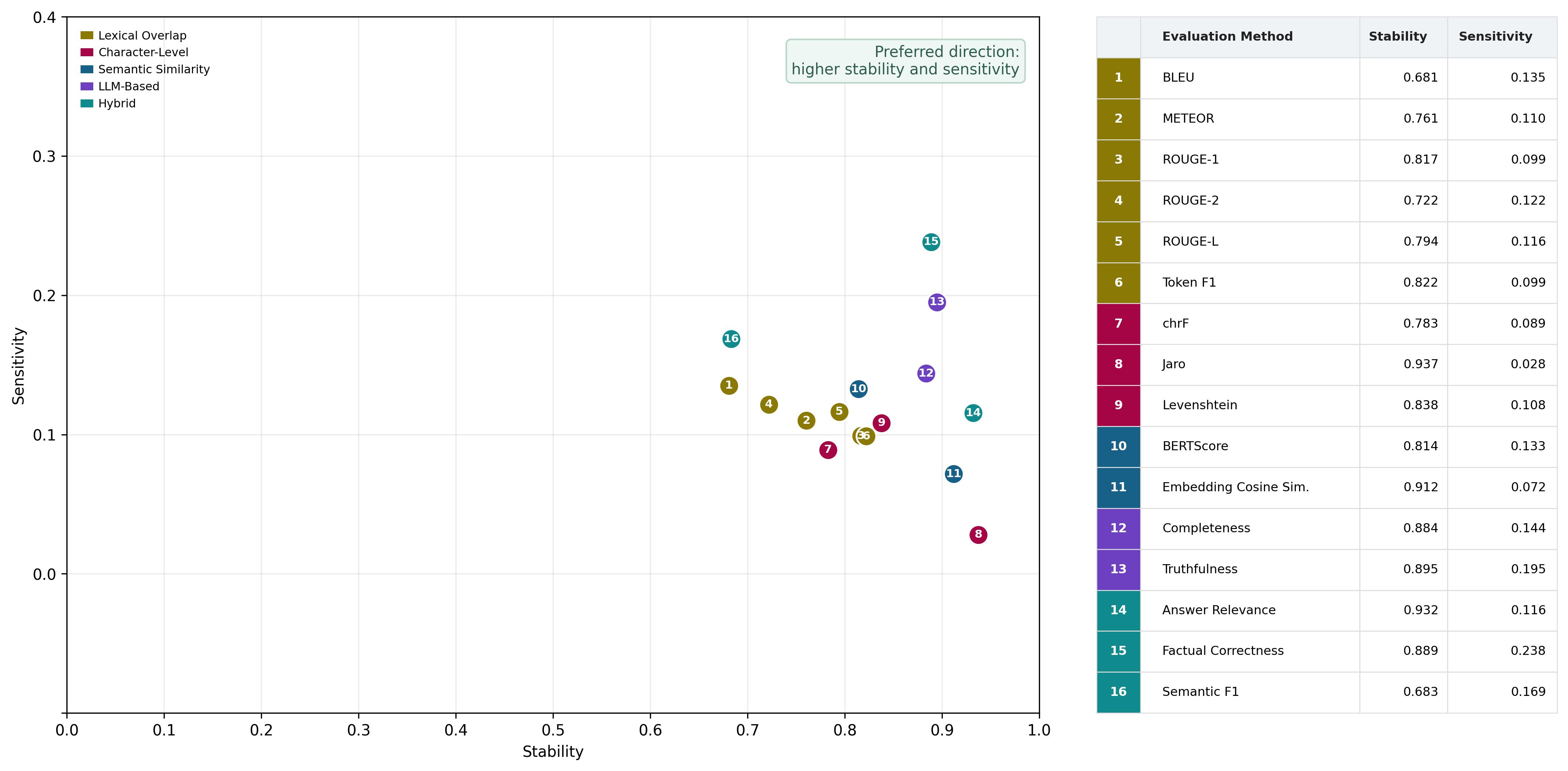}
    \caption{Stability--sensitivity characterization of automatic evaluation methods. Stability summarizes consistency under correctness-preserving transformations, while sensitivity summarizes responsiveness to correctness-altering transformations.}
    \label{fig:summary_level}
\end{figure}

Figure~\ref{fig:summary_level} summarizes evaluator behavior in terms of stability and sensitivity (Section~\ref{sec:analysis}). Because correctness-altering transformations introduce localized degradations rather than rendering responses entirely incorrect, sensitivity values are not expected to approach one.

The stability--sensitivity space reveals distinct behavioral regimes. Similarity-based methods generally exhibit moderate-to-high stability but relatively low sensitivity, indicating robustness to correctness-preserving transformations alongside weaker responses to correctness violations. Notably, Jaro achieves the highest stability (0.937) but the lowest sensitivity (0.028), illustrating that stability alone does not imply reliable detection of correctness violations.

LLM-based and hybrid evaluators exhibit more diverse profiles. Factual Correctness and Truthfulness provide the strongest combination of stability and sensitivity, reaching (0.889,0.238) and (0.895,0.195), respectively. Completeness and Answer Relevance achieve comparable stability but lower sensitivity, while Semantic F1 exhibits substantially lower stability despite relatively high sensitivity.

These observations demonstrate that evaluation methods within the same family are not behaviorally interchangeable. Although LLM-based evaluators are often grouped as ``LLM-as-a-Judge'' methods, they exhibit substantially different stability--sensitivity profiles. Consequently, evaluator selection should consider the correctness properties most important for the target application rather than the evaluation paradigm alone. More broadly, evaluators with similar aggregate performance may exhibit substantially different responses to correctness-preserving and correctness-altering transformations, highlighting behavioral differences obscured by aggregate scores.

\subsection{Repeat-run Variability}
\begin{figure}[!htbp]
    \centering
    \includegraphics[width=\linewidth]{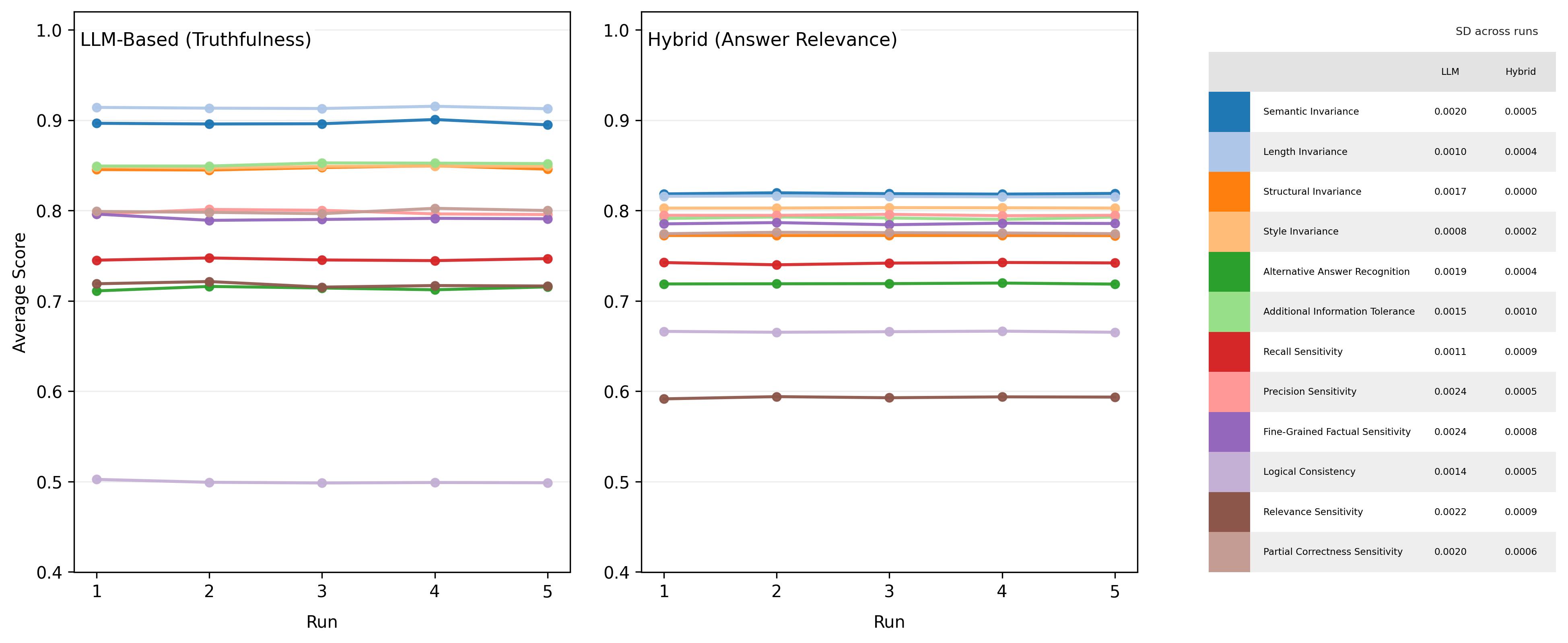}
    \caption{Assumption-level mean scores across five repeated evaluations for representative LLM-based (Truthfulness) and hybrid (Answer Relevance) evaluators. The accompanying table reports the standard deviation of the assumption-level scores across runs.}
    \label{fig:repeat_runs}
\end{figure}

LLM-based evaluators may produce different scores across repeated executions despite identical inputs. To assess the impact of this stochasticity, we repeat the complete evaluation procedure five times for a representative LLM-based evaluator (Truthfulness) and a representative hybrid evaluator (Answer Relevance) under identical settings. Figure~\ref{fig:repeat_runs} presents the mean assumption-level scores across the five repeated runs, together with the standard deviation of each assumption profile.

Both evaluators exhibit highly consistent assumption-level profiles across runs, with standard deviations below $0.0025$ for every correctness assumption. Answer Relevance shows particularly low variability, with all standard deviations below $0.001$, while Truthfulness exhibits only marginally larger fluctuations.
The relative ordering of assumption-level scores is also preserved across runs, demonstrating that the resulting evaluator characterization is highly reproducible despite stochastic inference.

\subsection{Configuration Sensitivity}
\begin{figure}[!htbp]
    \centering
    \includegraphics[width=\linewidth]{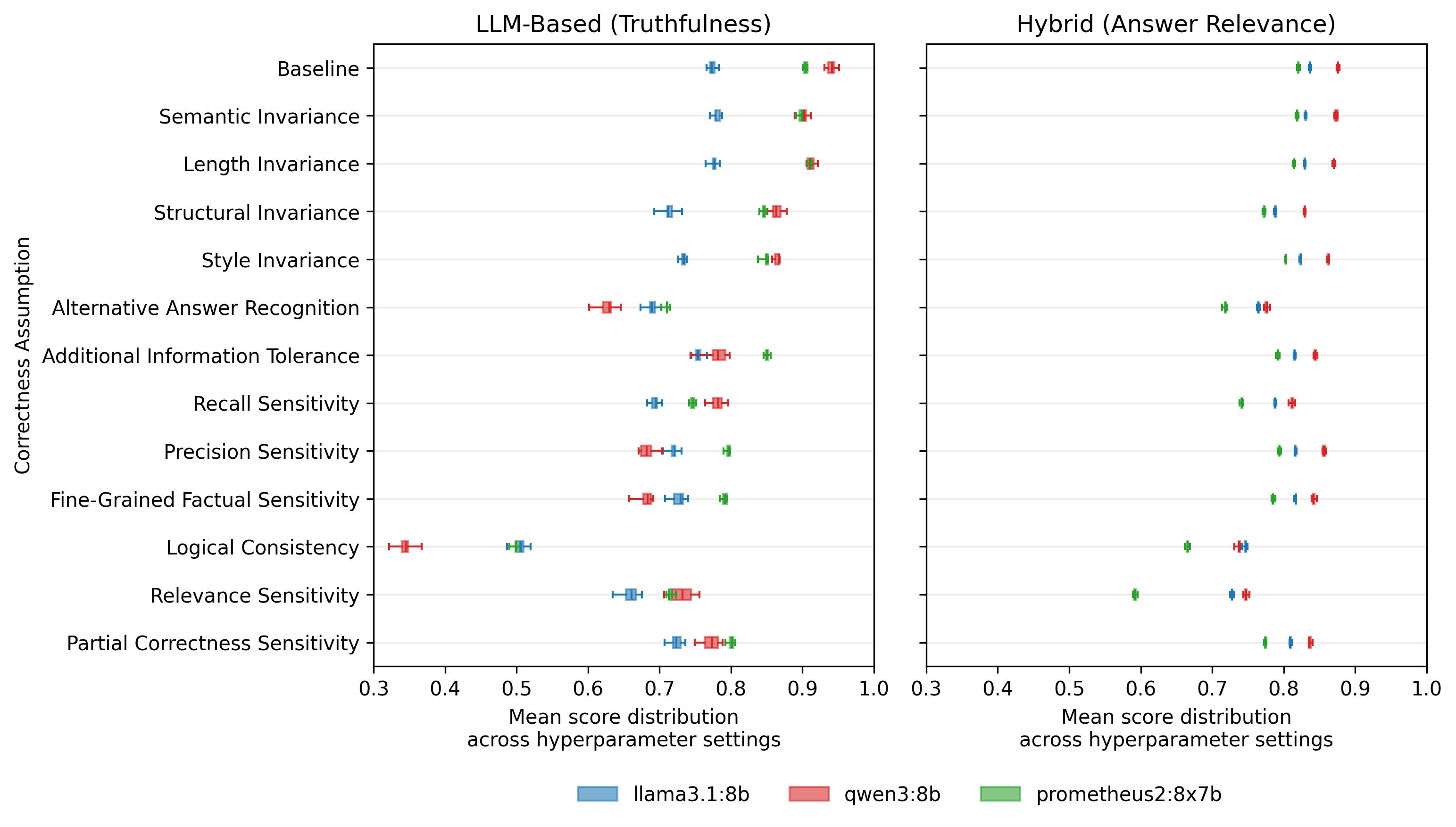}
    \caption{Assumption-level score distributions under different evaluator backbones and decoding configurations for representative LLM-based (Truthfulness) and hybrid (Answer Relevance) evaluators. Each marker denotes the mean score across decoding configurations for one backbone, with horizontal error bars indicating the standard deviation over all temperature, top-$p$, and top-$k$ settings.}
    \label{fig:configuration_sensitivity}
\end{figure}

Figure~\ref{fig:configuration_sensitivity} summarizes the distribution of assumption-level scores obtained across all combinations of decoding parameters and three evaluator backbones: Llama-3.1-8B, Qwen-3-8B, and Prometheus~2.

Overall, assumption-level profiles exhibit limited variation across decoding configurations within each backbone, as indicated by the relatively small error bars. In contrast, backbone choice produces substantially larger shifts in assumption-level scores across both correctness-preserving and correctness-altering assumptions.

The magnitude of these shifts varies by evaluator and assumption. Truthfulness exhibits pronounced backbone differences across several assumptions, whereas Answer Relevance generally shows smaller differences, although notable backbone effects remain for some assumptions. Overall, backbone choice has a larger influence on the observed evaluator profiles than the decoding variations considered here, highlighting the importance of reporting the evaluator backbone.

\subsection{Test Suite Reproducibility}

\begin{figure}[!htbp]
    \centering
    \includegraphics[width=\linewidth]{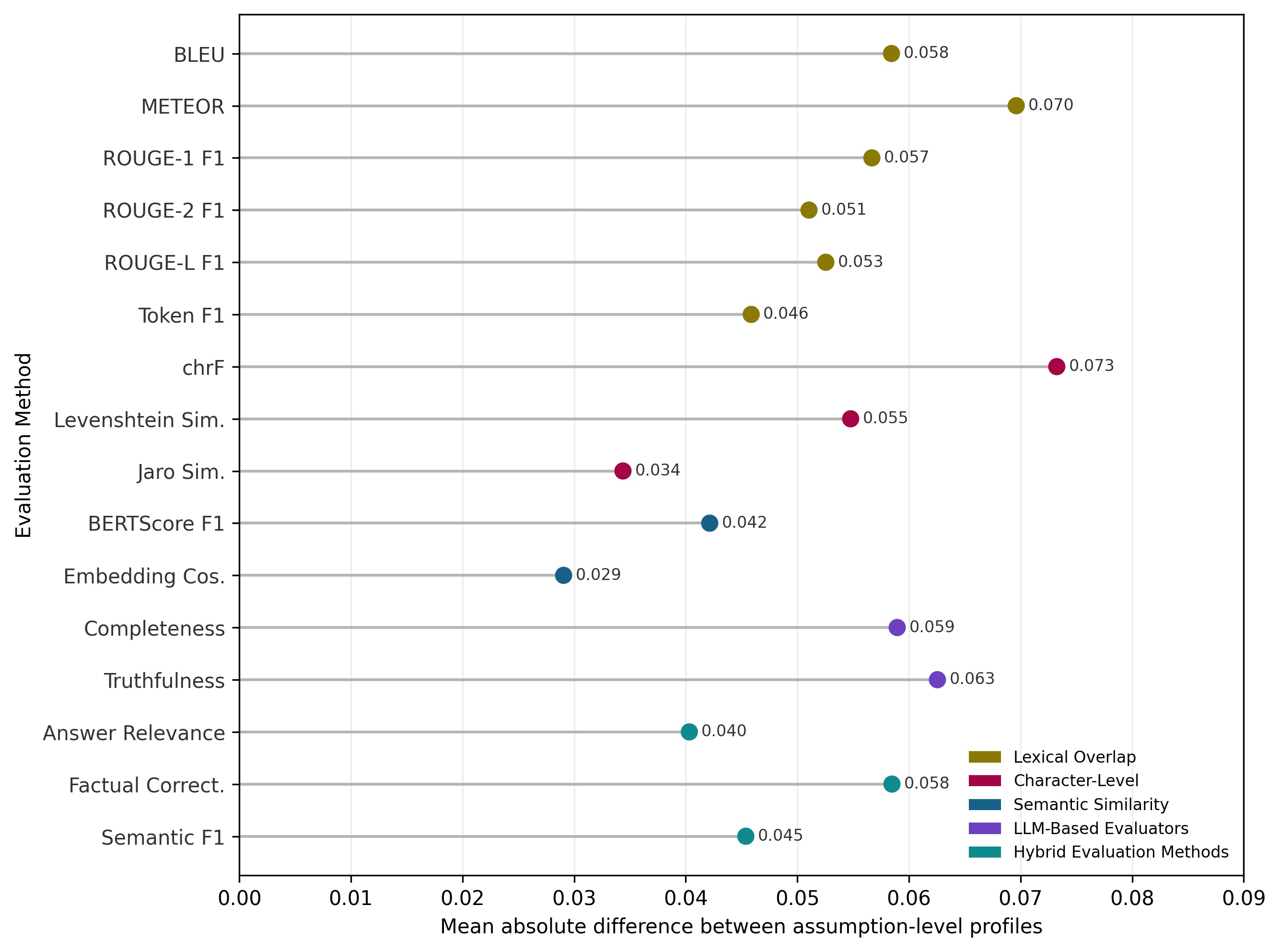}
    \caption{Reproducibility of assumption-level evaluator profiles across independently generated evaluation test suites. Each point represents the mean absolute difference (MAE) between the assumption-level score differences obtained from the original and regenerated test suites for one evaluation method.}
    \label{fig:testsuite_reproducibility}
\end{figure}

Figure~\ref{fig:testsuite_reproducibility} summarizes the reproducibility of assumption-level evaluator profiles when the test suite is independently regenerated using Qwen-3-8B.

All evaluators obtain profile MAE values below $0.075$, indicating modest differences between the original and regenerated test suites. This suggests that the resulting profiles capture systematic evaluator behavior rather than being dominated by a particular realization of the generated test suite. These values should be interpreted as measures of profile consistency rather than evaluator quality, since an evaluator that responds weakly across both test suites may also obtain a low MAE.

The degree of reproducibility nevertheless varies across evaluation methods. Embedding Cosine Similarity and Jaro Similarity exhibit the smallest profile differences, indicating that their assumption-level characterizations are least affected by regeneration. Conversely, METEOR and chrF produce the largest MAE values, suggesting that their behavioral profiles are comparatively more sensitive to variation in the generated baselines and transformations. The remaining evaluators occupy a relatively narrow intermediate range, with no clear separation by evaluator family. The regenerated stability--sensitivity characterization is provided in Appendix \ref{app:results_reprod}, showing that the principal behavioral patterns are broadly preserved.

Overall, the low MAE values and preservation of the main stability--sensitivity patterns indicate that test-suite regeneration changes the magnitude of some measured effects without materially altering the qualitative behavioral characterization of the evaluators.

\section{Discussion}

\subsection{Practical Implications}
The proposed framework highlights a limitation of conventional benchmark evaluation, which typically summarizes evaluator quality using correlation with human judgments or an overall benchmark score. Our results reveal that evaluation methods with similar aggregate performance may nevertheless respond very differently to different aspects of response quality.

The findings further suggest that automatic evaluator selection should be guided by the intended evaluation objective rather than aggregate benchmark performance alone. Applications requiring robustness to semantically equivalent reformulations, such as paraphrasing or stylistic variation, benefit from evaluators exhibiting high stability while remaining responsive to genuine correctness violations. Conversely, applications in which factual reliability is paramount require evaluators that consistently detect omissions, factual inaccuracies, and logical inconsistencies without being overly sensitive to superficial response variations. Since no evaluation method simultaneously satisfies all proposed correctness assumptions, automatic evaluation should be viewed as a trade-off between complementary behavioral properties rather than a search for a universally optimal metric.

The analysis also reveals substantial variation within evaluator families. LLM-based and hybrid evaluators exhibit distinct behavioral profiles across correctness assumptions, while similarity-based methods also differ in their stability and sensitivity to particular response transformations. Consequently, categorizing evaluation methods solely as lexical, semantic, hybrid, or LLM-based provides only a coarse characterization of their behavior.

The principal contribution of the proposed framework is not another automatic evaluation metric, but a methodology for systematically characterizing existing evaluation methods. Rather than replacing conventional benchmark evaluation, it provides a complementary behavioral perspective that identifies the correctness properties under which evaluators succeed or fail. This diagnostic perspective can also facilitate more informed evaluator selection and support the development of future automatic evaluation methods.

\subsection{Limitations}

Although the proposed framework is broadly applicable, several limitations remain and motivate future research.

First, the evaluation test suite relies on LLM-generated baseline responses and controlled transformations. Although the robustness analyses demonstrate that the resulting behavioral characterizations are largely robust to stochastic generation and generator choice, automatically generated transformations cannot capture the full diversity of naturally occurring response variations. Incorporating human-authored transformations or alternative generation strategies could further strengthen the framework.

Second, the proposed correctness assumptions are not exhaustive. Other dimensions of evaluator behavior, including reasoning quality, uncertainty calibration, safety, and domain-specific requirements, are not explicitly modeled. The framework could therefore be extended with assumptions tailored to additional evaluation objectives and application domains.

Third, our experiments focus on question answering over a fixed collection of documents and reference answers. Although the methodology is largely task-agnostic, applying it to other generation tasks, such as summarization, dialogue, long-form generation, or code generation, may require task-specific correctness assumptions and transformation procedures. Its generality across diverse generation tasks therefore remains to be established.

Finally, the current framework evaluates correctness relative to a single reference answer, reflecting the design of many existing evaluation benchmarks. However, open-ended generation tasks often admit multiple equally valid responses that differ substantially in wording, structure, or content organization. Extending the framework to multi-reference settings would enable a broader characterization of evaluator behavior under more diverse notions of acceptable correctness.

\section{Conclusion}
We introduced a diagnostic framework for characterizing reference-based automatic evaluation methods through controlled response transformations. We defined correctness-preserving and correctness-altering assumptions and evaluated lexical, semantic, hybrid, and LLM-based evaluators at transformation, assumption, and summary levels.
Our experiments reveal distinct behavioral profiles across evaluation methods, with no evaluator satisfying all proposed correctness assumptions. Evaluators exhibit different trade-offs between stability under correctness-preserving transformations and sensitivity to correctness violations, with substantial variation even within evaluator families. Robustness analyses further show highly consistent profiles across repeated runs and limited variation across decoding configurations within a given backbone, while backbone choice produces larger shifts. Independently regenerating the test suite also changes the magnitude of some effects while broadly preserving the main behavioral characterization.
These findings establish behavioral correctness assumptions as a complementary methodology for systematically characterizing the strengths and limitations of automatic evaluation methods and provide a foundation for developing evaluators with more explicit and testable behavioral properties.

\section*{Funding Information}
This research was funded under Interagency Agreement 70RSAT23KPM000049 by the U.S. Department of Homeland Security (DHS) Science and Technology (S\&T) Directorate.

\bibliographystyle{acm}
\bibliography{ref}

\onecolumn

\appendix

\section{Additional Evaluation Method Details}
\label{app:llm_config}

\begin{figure}[!htbp]
    \centering
    \includegraphics[width=.9\linewidth]{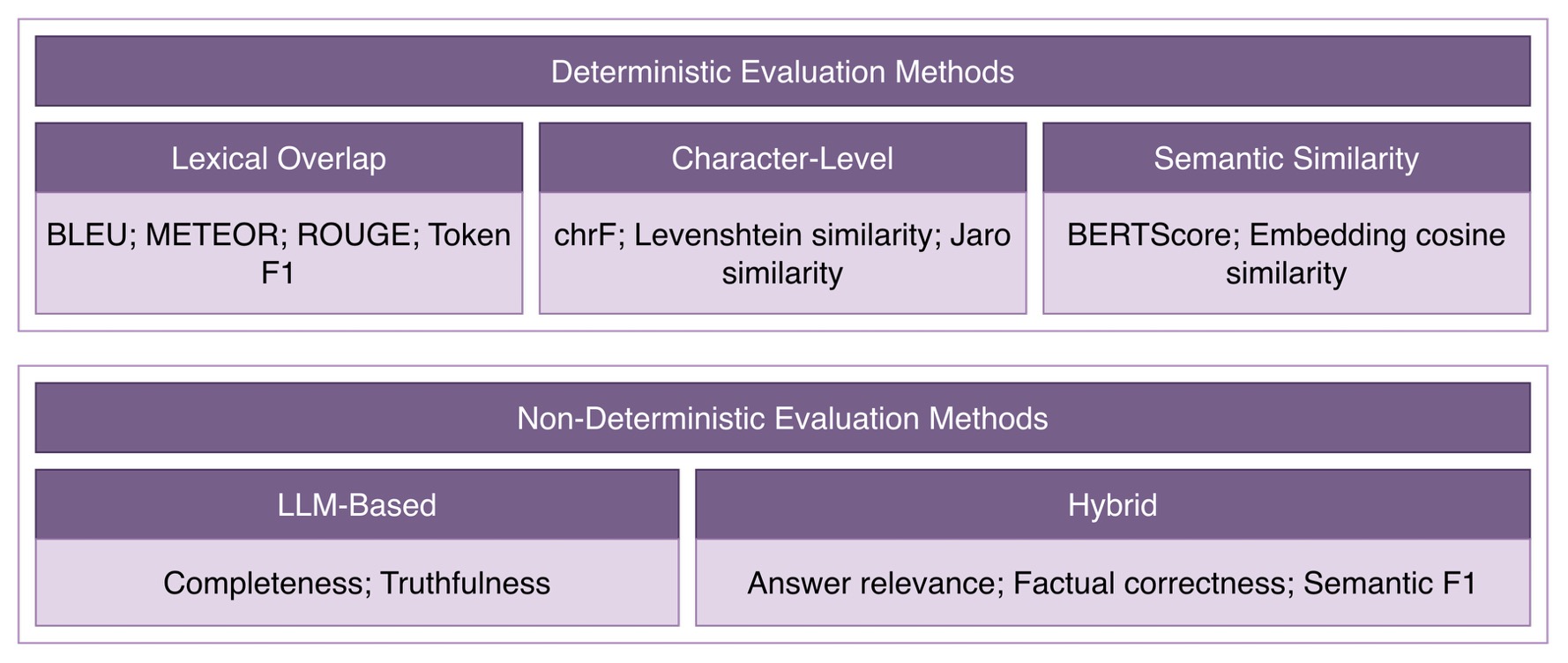}
    \caption{Taxonomy of reference-based automatic evaluation methods included in this study.}
    \label{fig:evaluation_methods}
\end{figure}

Figure~\ref{fig:evaluation_methods} summarizes the evaluation methods included in this study. We group these methods into five categories according to their underlying evaluation paradigm: lexical overlap, character-level, semantic similarity, LLM-based, and hybrid evaluation methods. We consider commonly used lexical-overlap metrics (BLEU, METEOR, ROUGE-1, ROUGE-2, ROUGE-L, and Token F1), character-level metrics (chrF, Jaro Similarity, and Levenshtein Similarity), and semantic-similarity metrics (BERTScore and Embedding Cosine Similarity).

For LLM-based evaluators, we use Truthfulness and Completeness from RAG Playground \cite{papadimitriou2024rag}. Truthfulness uses an LLM to assess the factual accuracy of the generated response with respect to the reference, considering both consistency and contradiction. Completeness uses an LLM to assess whether the generated response covers the key information in the reference, considering both explicit and implicit information and identifying missing important content.

Hybrid methods combine LLM-based evaluation with deterministic components such as embedding-based similarity or structured score computation. We consider Answer Relevance and Semantic F1 from RAG Playground \cite{papadimitriou2024rag}, and Factual Correctness from RAGAS \cite{es2024ragas}. Answer Relevance combines embedding similarity with an LLM assessment of semantic meaning and relevance, averaging the two components with equal weight to obtain the final score. Semantic F1 uses embedding-based matching of key points between the generated response and reference, treating pairs above a predefined similarity threshold as matches. Precision and recall are then computed from the matched key points and combined into an F1 score. Factual Correctness decomposes the response and reference into claims using an LLM, applies natural language inference to determine claim-level factual overlap, and quantifies this overlap using F1.

Lexical overlap, character-level, and semantic similarity methods are deterministic, producing identical scores for repeated evaluations of the same response-reference pair. In contrast, LLM-based and hybrid evaluation methods rely wholly or partially on LLMs for scoring and may therefore exhibit non-deterministic behavior.
This categorization enables our framework to assess evaluation methods across diverse evaluation paradigms while remaining independent of the specific implementation of any individual method.

For LLM-based and Hybrid evaluation methods, we use the Prometheus-2 (8$\times$7B) \cite{kim2024prometheus} model.
Prometheus-2 is an open-source language model specifically fine-tuned for LLM-as-a-Judge tasks, including absolute grading and pairwise ranking. We use the 4-bit quantized version released by TensorTemplar and execute it locally using the Ollama runtime. Llama-3.1-8B \cite{grattafiori2024llama} and Qwen-3-8B \cite{yang2025qwen3} are included as strong open-source general-purpose language models to evaluate whether the observed correctness assumption profiles are specific to a judge-specialized model or remain consistent across different LLM backbones.
Unless otherwise specified, all experiments reported in this paper use Prometheus-2 (8$\times$7B) as the LLM-based evaluator with decoding configuration fixed at $\text{temperature}=0.2$, $\text{top-p}=0.8$, and $\text{top-k}=40$.

\section{Prompt design}
\label{app:prompt}
\subsection{Reference Answer}
\label{app:prompt_reference}
The LLM was prompted carefully to generate desirable question types. The questions were designed to be realistic and detailed, representative of items that an individual who is well-informed of role responsibilities and general organizational operations would reasonably query. Strict instructions were provided to the LLM to only ask questions that could be answered using the provided information. As such, in addition to ``question'' and ``reference answer'' fields in the generated output, ``supporting text'' that corresponded to the text from which the answer could be derived was also included. The LLM was hosted with a vLLM server.
A small random sample of the questions and corresponding reference answers was reviewed by a subject-matter expert to assess their quality.
The prompt used for reference answer generation is provided in Table~\ref{tab:prompt_list}.

\subsection{Transformation}
\label{app:prompt_transformation}
Each transformation is produced using an LLM guided by transformation-specific prompting instructions.
The LLM is also provided with the supporting context that was used to generate the reference question-answer pair.
The prompts explicitly specify the desired modification while constraining the model to avoid introducing additional changes beyond those required for the target transformation. For example, the `paraphrase' transformation rewrites the response while preserving its meaning, whereas a `incorrect fact added' transformation introduces an incorrect factual statement while leaving the remaining response unchanged.
Table~\ref{tab:prompt_list} summarizes the prompting templates used to generate the complete evaluation test suite.
For each baseline response, one transformed response is generated for every transformation listed in Table~\ref{tab:correctness_assumptions_perturbations}. Collectively, these transformed responses constitute a controlled evaluation test suite in which each evaluation instance isolates a single correctness assumption, enabling systematic analysis of evaluator behavior across diverse response modifications.
Baseline responses and controlled response transformations were generated using the Llama-3.1-8B model \cite{grattafiori2024llama}.
A decoding configuration of $\text{temperature}=0.8$, $\text{top-p}=0.9$, and $\text{top-k}=40$ was used to encourage diverse yet coherent response generation.

\begin{table}[!htbp]
\centering
\caption{LLM prompts used to generate the evaluation test suite.}
\label{tab:prompt_list}
\begin{tabular}{p{16cm}}
\toprule
\textbf{Stage: Prompt} \\

\midrule

Reference: You are a highly trained and informed U.S. Coast Guard officer stationed in Boston, responsible for answering questions about the local area, maritime regulations, and safety procedures. Use the given information to generate scenario-based exam questions and answers for junior officers to test their ability to respond appropriately to incidents. Questions should be like ``what would you do if...'', ``what is an appropriate response to X...''. They should NOT be simple, factual questions about general Coast Guard operations. 

ONLY ask questions that can be answered using the provided information. Do not include any information that is not directly supported by the text. The answers should be detailed and contain 5-8 sentences.

supporting text should be a direct quote from the provided information that supports the answer. do not shortcut or paraphrase, paste the entirety of the relevant text as the supporting text. no ``...'' should be included in the supporting text.

Generate 5 training examples from each of the documents in the following JSON structure:

    [
    {{
      question: $<...>$, 
      
      answer: $<$5-8 sentences response to the question. do not include citations or references to specific documents in the answer$>$, 
      
      supporting text: $<$direct quote from the document that supports the answer$>$
    }}
  ]

  \\ \hline

Baseline: Rewrite the ground truth as a natural answer a RAG system would produce. The rewritten version should preserve the same meaning but use a different sentence structure. \\ \hline

Paraphrase: Rewrite the assumed response in a slightly different phrasing, possibly reordering information, using different wording, synonyms, and a different grammatical structure. \\ \hline

Verbose: Expand the assumed response into a fuller paragraph with more explanatory wording. For each fact in the assumed response, add a short clarification of what it means or why it matters. Preserve every fact from the assumed response, but do not introduce any new factual claims. Do not copy sentences from the supporting context. Do not add new requirements, entities, conditions, numbers, or procedures that are not already in the assumed response. The output should be longer than the assumed response while remaining faithful to it. \\ \hline

Concise: Make the assumed response shorter than the assumed response without changing factual content. \\ \hline

Structure Change: Change the format of the assumed response, such as prose to bullets, while preserving facts. \\ \hline

Uncertain Style: Rewrite the assumed response in an uncertain tone without changing factual content. \\ \hline

Certain Style: Rewrite the assumed response in a certain tone without changing factual content.  \\ \hline

Alternative Correct Answer: Write a different answer that is still fully correct for the question. \\ \hline

Correct Fact Added: In the assumed response, add one correct, relevant fact without introducing unsupported information. \\ \hline

Key Detail Omitted: Remove one key piece of information so the answer becomes incomplete. \\ \hline

Incorrect Fact Added: In the assumed response, add one plausible but false or unsupported statement while keeping the original answer mostly intact. \\ \hline

Subtle Factual Error: Introduce a small factual error, such as changing an entity, number, condition, or relationship. \\ \hline

Logical Contradiction: Rewrite the answer so it contains an internal contradiction. \\ \hline

Irrelevant Sentence Added: In the assumed response, add one irrelevant off-topic sentence without changing the original facts.  \\ \hline

Partially Correct: Keep part of the answer correct, but make part incorrect or missing.  \\
\bottomrule
\end{tabular}
\end{table}

\section{Data Statistics}
\label{app:data_statistics}
Table~\ref{tab:testsuite_stats} summarizes the characteristics of the evaluation test suite. On average, reference answers contain 55.09 words, while the corresponding baseline responses contain 34.31 words, reflecting the natural rewriting process used to generate evaluation responses. The controlled transformations exhibit the expected variation in response length, with verbose transformations substantially increasing response length, concise and key-detail-omitted transformations producing shorter responses, and most remaining transformations preserving lengths comparable to the baseline.

\begin{table}[!htbp]
\centering
\small
\caption{Statistics of the evaluation test suite.}
\label{tab:testsuite_stats}
\begin{tabular}{lr}
\toprule
\textbf{Statistic} & \textbf{Value} \\
\midrule
Source documents & 9 \\
Total question--answer pairs & 203 \\
Average QA pairs per document & 22.55 \\
Average reference answer length & 55.09 words \\
Average baseline response length & 34.31 words \\
\midrule
\multicolumn{2}{c}{\textbf{Average response length by transformation (words)}} \\
\midrule
Paraphrase & 36.93 \\
Verbose & 124.47 \\
Concise & 25.40 \\
Structure Change & 34.46 \\
Certain Style & 32.62 \\
Uncertain Style & 46.38 \\
Alternative Correct Answer & 31.06 \\
Correct Fact Added & 44.44 \\
Key Detail Omitted  & 22.34 \\
Incorrect Fact Added & 49.48 \\
Subtle Factual Error & 37.48 \\
Logical Contradiction & 39.48 \\
Irrelevant Sentence Added & 43.03 \\
Partially Correct & 31.58 \\
\midrule
Total transformed responses & 2842 \\
Total evaluation instances & 3045 \\
\bottomrule
\end{tabular}
\end{table}

\section{Additional Results}
\label{app:results}

\subsection{Absolute Transformation-Level Scores}
\label{app:results_trans}
\begin{figure}[!htbp]
    \centering
    \includegraphics[width=\linewidth]{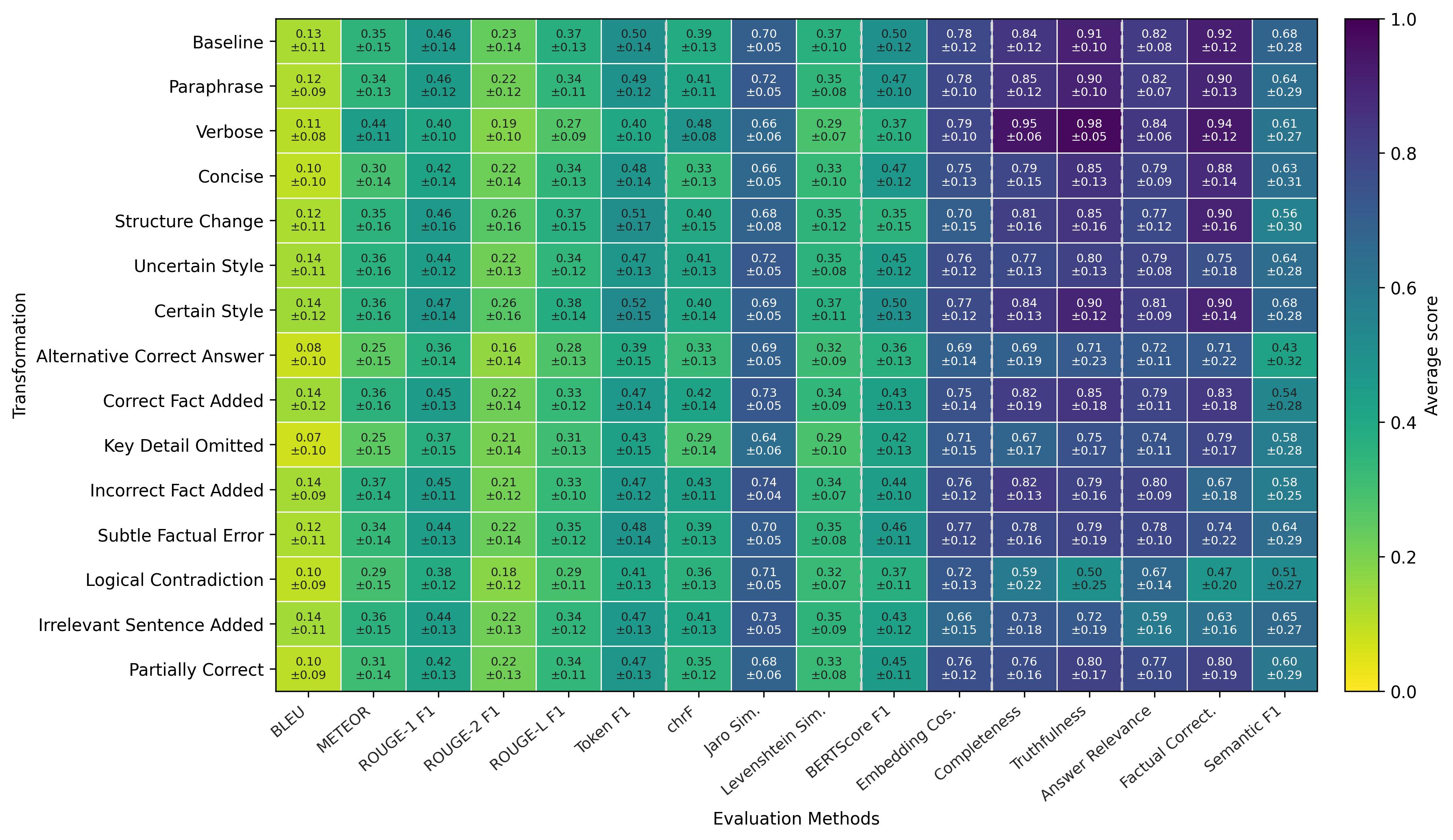}
    \caption{Average score for the baseline response and for each controlled transformation and evaluation method.}
    \label{fig:transformation_absolute}
\end{figure}

Figure~\ref{fig:transformation_absolute} presents the absolute scores for each controlled transformation, complementing the baseline-relative differences reported in Figure~\ref{fig:transformation_level}. Small relative differences do not necessarily indicate desirable behavior, as some evaluators assign relatively low scores to both the baseline and transformed responses. The absolute scores therefore provide additional context for interpreting the transformation-level results.

\subsection{Stability--Sensitivity Analysis after Test-Suite Regeneration}
\label{app:results_reprod}
\begin{figure}[!htbp]
    \centering
    \includegraphics[width=\linewidth]{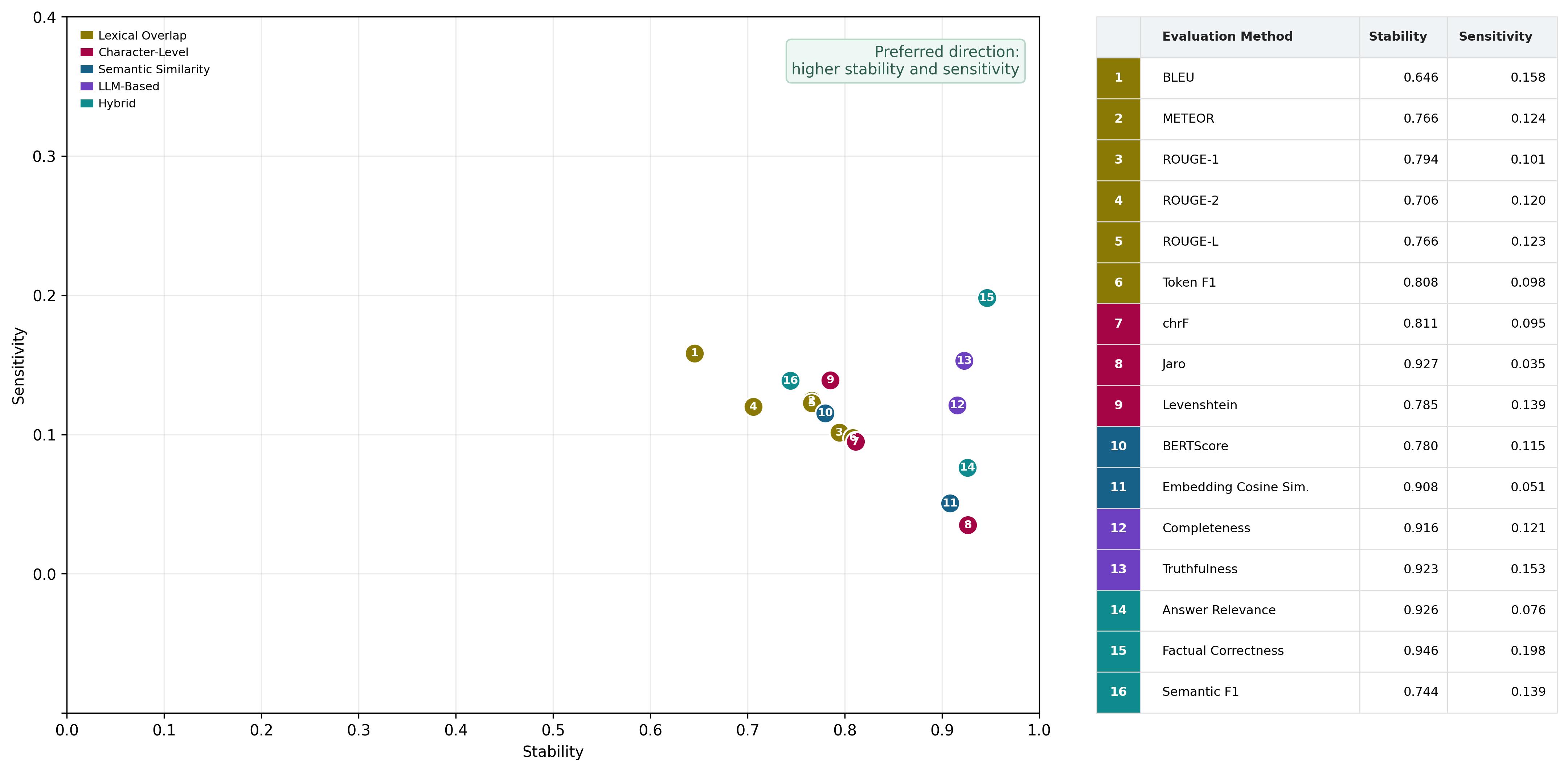}
    \caption{Stability--sensitivity characterization of evaluation methods obtained using the independently regenerated evaluation test suite. The regenerated benchmark is constructed using a different language model while preserving the same retrieval context, prompting strategy, and controlled transformation procedure.}
    \label{fig:testsuite_reproducibility_summary}
\end{figure}

Figure~\ref{fig:testsuite_reproducibility_summary} presents the stability--sensitivity characterization obtained from the regenerated test suite. The broad organization of the evaluator space is preserved despite shifts in individual stability and sensitivity values. Factual Correctness remains the most sensitive evaluator while maintaining high stability, and Truthfulness also retains relatively high stability and sensitivity. Jaro and Embedding Cosine Similarity remain highly stable but comparatively insensitive to correctness-altering transformations. Answer Relevance maintains high stability, although its sensitivity decreases on the regenerated test suite. Thus, test-suite regeneration affects the magnitude of the measured effects more than the qualitative characterization of evaluator behavior.

\end{document}